\documentclass{article}
\usepackage{ijcai25}
\usepackage{xspace}
\usepackage{times}
\usepackage{soul}
\usepackage{url}
\usepackage[hidelinks]{hyperref}
\usepackage[utf8]{inputenc}
\usepackage[small]{caption}
\usepackage{graphicx}
\usepackage{amsmath}
\usepackage{amsthm}
\usepackage{booktabs}
\usepackage{algorithm}
\usepackage{algorithmic}
\usepackage[switch]{lineno}
\usepackage{multirow}
\usepackage{color, colortbl, xcolor}
\usepackage{makecell}
\usepackage{subcaption}
\usepackage{amssymb}

\usepackage[symbol]{footmisc}

\newcommand{\modelname}{DFPG\xspace}

\definecolor{cvprblue}{rgb}{0.21,0.49,0.74}
\definecolor{qqb_blue}{RGB}{103,171,214}
\newcommand{\qqb}[1]{\textcolor{black}{#1}}

\def\ie{\textit{i.e}.,\xspace}
\def\eg{\textit{e.g}.,\xspace}

\newcommand{\yhc}[1]{{\color{black}#1}}
\definecolor{gold}{RGB}{222,174,0}

\newcommand{\dcl}[1]{{\color{black}#1}}

\title{Dual-level Fuzzy Learning with Patch Guidance for Image Ordinal Regression}

\author{
Chunlai~Dong$^{1,3,4,7}$\and
Haochao~Ying$^{2,3,7,}$\thanks{Corresponding Author: Haochao Ying and Jian Wu}\and
Qibo~Qiu$^{5}$\and
Jinhong~Wang$^{1,3,4,7}$\and
Danny~Chen$^{6}$\and
Jian~Wu$^{2,3,4,7,}$\footnotemark[1] \\ 
\affiliations
$^1$College of Computer Science and Technology, Zhejiang University\\
$^2$School of Public Health, Zhejiang University\\
$^3$State Key Laboratory of Transvascular Implantation Devices, \\The Second Affiliated Hospital Zhejiang University School of Medicine\\
$^4$Zhejiang Key Laboratory of Medical Imaging Artificial Intelligence\\
$^5$China Mobile (Zhejiang) Research \& Innovation Institute\\
$^6$College of Engineering, University of Notre Dame\\
$^7$Transvascular Implantation Devices Research Institute\\
\emails
\{dongcl, haochaoying, qiuqibo\_zju, wangjinhong, wujian2000\}@zju.edu.cn,
dchen@nd.edu
}

\begin{document}

\maketitle

\begin{abstract}
Ordinal regression bridges regression and classification by assigning objects to ordered classes. While human experts rely on discriminative patch-level features for decisions, current approaches are limited by the availability of only image-level ordinal labels, overlooking fine-grained patch-level characteristics.
\dcl{In this paper, we propose a \underline{D}ual-level \underline{F}uzzy Learning with \underline{P}atch \underline{G}uidance framework, named \modelname that learns precise feature-based grading boundaries from ambiguous ordinal labels, with patch-level supervision.}
\dcl{Specifically, we propose patch-labeling and filtering strategies to enable the model to focus on patch-level features exclusively with only image-level ordinal labels available.}
\dcl{
We further design a dual-level fuzzy learning module, which leverages fuzzy logic to quantitatively capture and handle label ambiguity from both patch-wise and channel-wise perspectives.}
Extensive experiments on various image ordinal regression datasets demonstrate the superiority of our proposed method, further confirming its 
ability in distinguishing samples from \dcl{difficult-to-classify }categories. The code is available at 
\href{https://github.com/ZJUMAI/DFPG-ord}{https://github.com/ZJUMAI/DFPG-ord}.
\end{abstract}

\section{Introduction}
\label{sec:intro}

\yhc{Image ordinal regression, also known as ordinal classification in computer vision, aims to infer the ordinal labels of images. This task resides at a crucial intersection of the fundamental classification and regression paradigms, where class labels display inherent sequential or logical relationships. The image ordinal regression (grading) methodology has exhibited significant utility across diverse computer vision applications, including facial age estimation~\cite{Li_2019_CVPR,10.1007/978-3-030-58592-1_23}, image aesthetic assessment~\cite{10.1007/978-3-319-46448-0_40,Lee2019ImageAA}, historical image dating~\cite{hci_1}, and medical disease grading~\cite{10294274}}.

\yhc{Known ordinal regression approaches predominantly followed either regression or classification paradigms~\cite{10.1007/978-3-319-16811-1_10,7406390}, employing conventional optimization objectives such as mean absolute/square error, or cross-entropy loss.}
However, these methods did not fully leverage the unique characteristics of ordinal regression, \ie the labels 
exhibit both ambiguity and ordinality.
Specifically, for example, in disease grading, the boundary between scores 1 and 2 is ambiguous, and the final value often relies on a doctor's subjective judgment. Meanwhile, the ordinality reflects that a higher score indicates a more severe disease.
\yhc{To address the ambiguity,}
several studies converted the labels of samples into a label distribution, leveraging the robustness of the distribution to deal with the inherent ambiguity among ordinal labels~\cite{8578652,7890384}. 
\qqb{When exploiting the ordinality,} some classification-based approaches treated ordinal regression as a multi-class classification task, developing new strategies to assist in learning inter-class ordinal relationships (e.g., soft labeling~\cite{softlabel} or binary label sequence prediction~\cite{ord2seq}).
Some ranking-based methods compared samples with specific anchors to learn the ordinal relationships~\cite{Lee2021DeepRC,MWR}. 

\begin{figure}[tbp]
	\centering
	\begin{subfigure}{0.45\linewidth}
		\centering
		\includegraphics[width=0.85\linewidth]{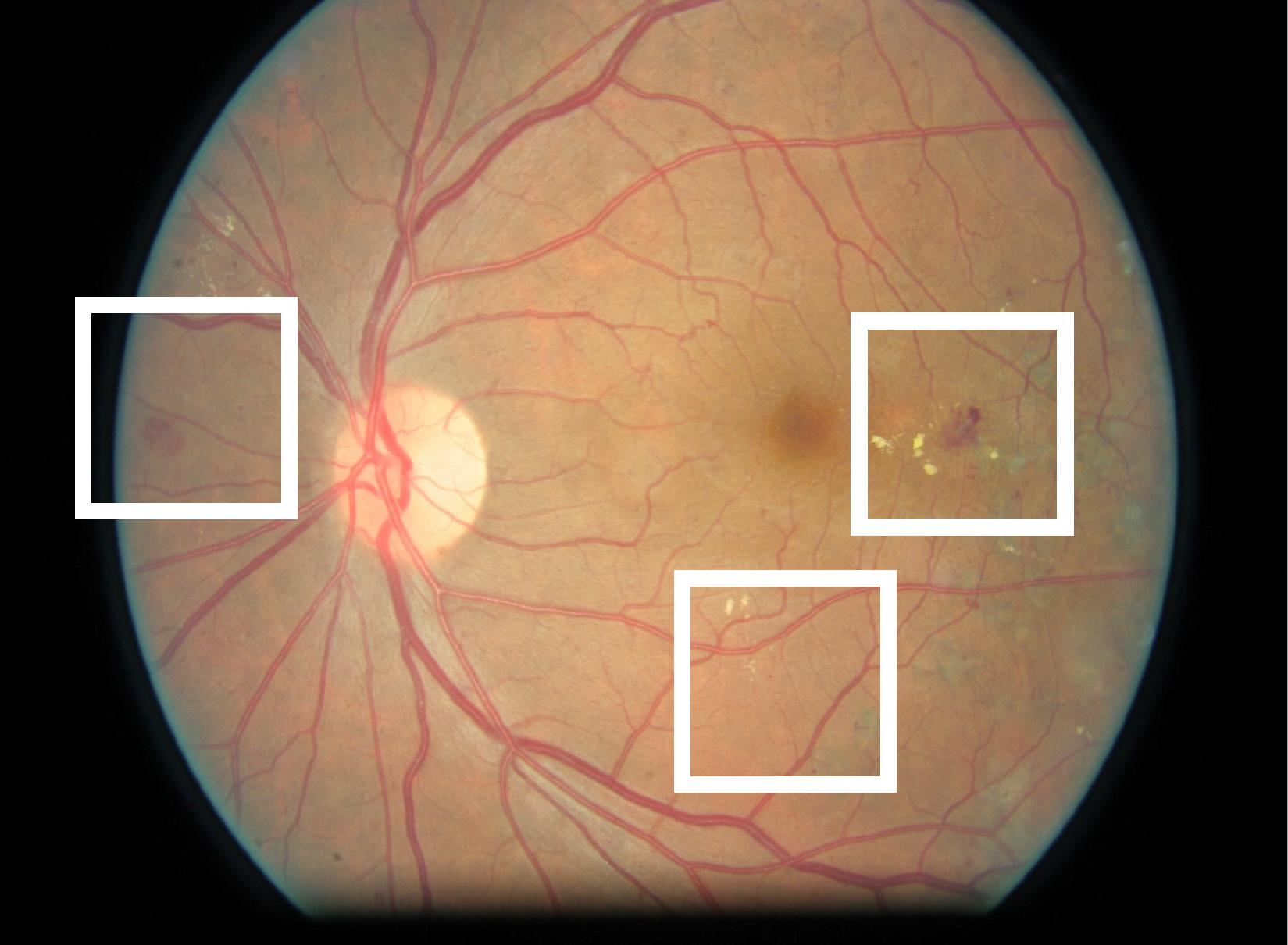}
        \vspace{-1ex}
		\caption{}
        \vspace{-1ex}
		\label{chutian3}
	\end{subfigure}
	\centering
	\begin{subfigure}{0.45\linewidth}
		\centering
		\includegraphics[width=0.85\linewidth]{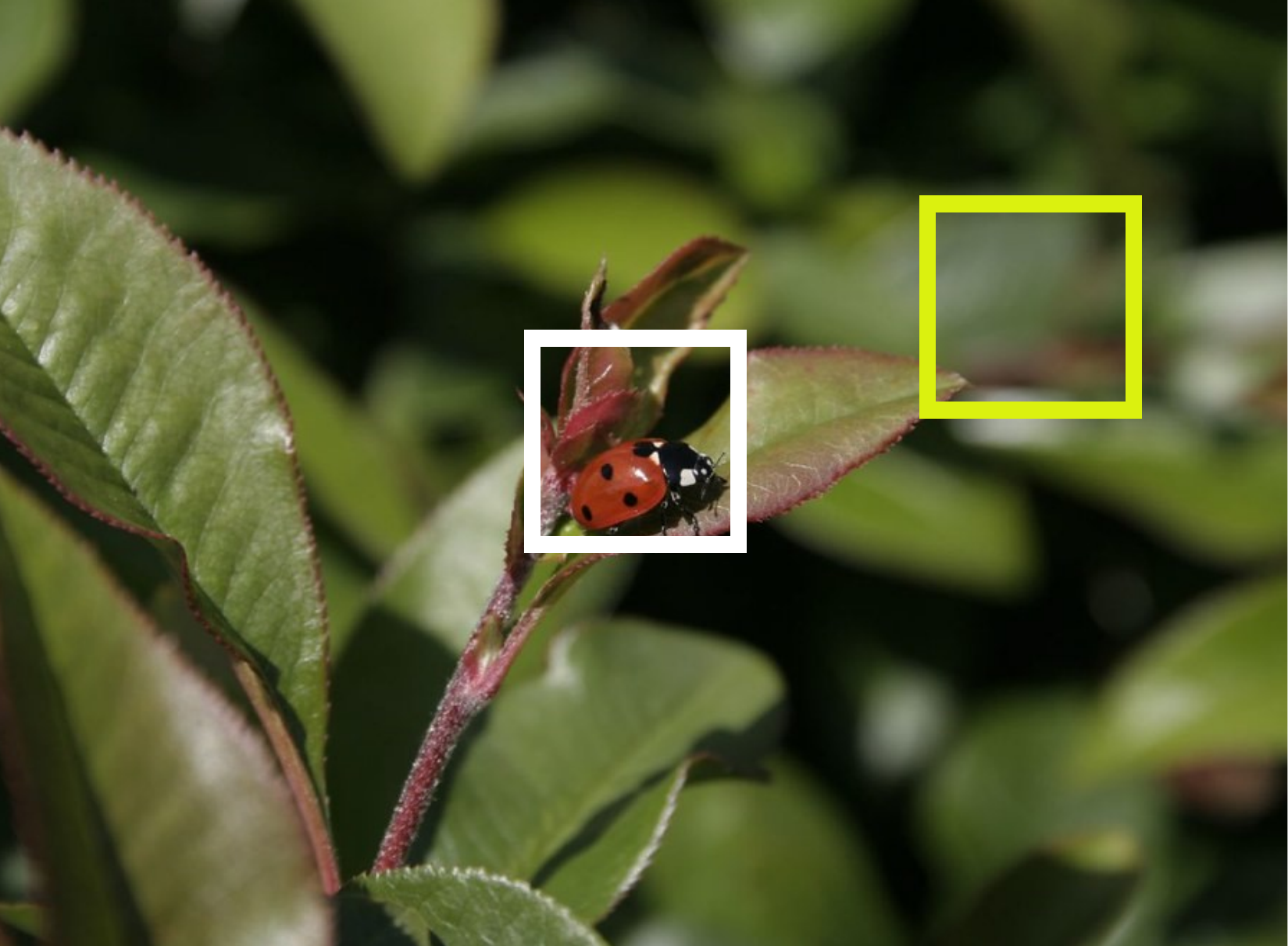}
        \vspace{-1ex}
		\caption{}
        \vspace{-1ex}
		\label{chutian3}
	\end{subfigure}
    \vspace{-1ex}
    \caption{\yhc{Illustrating} some influencing factors in human decision-making processes. (a) Doctors focus on specific lesion areas in the DR grading scenario. 
  (b) Evaluators may increase aesthetic scores based on certain regions (\eg white rectangle) while reducing scores due to blurred areas (\eg yellow rectangle).}
  \label{fig:region_decision} 
  \vspace{-3ex}
\end{figure}

\qqb{Despite the effectiveness of the aforementioned methods in improving the performance of ordinal regression~\cite{MWR,ord2seq}, they neglected an essential phenomenon: When making image grading decisions, it is actually the discriminative patch-level features that guide the human decision-making process.}
For instance, \qqb{as shown in Figure~\ref{fig:region_decision},} in the context of Diabetic Retinopathy (DR) grading, \yhc{clinicians determine} the severity of the disease by focusing only on specific DR lesion features within a small portion of the fundus image regions, such as hemorrhages and soft exudates~\cite{10596287}. \qqb{Similarly,} in the context of image aesthetic assessment, evaluators often base their scores not only on the overall information of an image but also tend to raise their ratings because specific regions align with their personal aesthetic preferences~\cite{annurev:/content/journals/10.1146/annurev-psych-120710-100504}. 
\yhc{Hence, a key challenge to curent research is: How to effectively exploit patch-level granular features within the ordinal regression framework, especially given the constraint of only image-level ordinal labels available.}

\yhc{In this paper, we propose a \underline{D}ual-level \underline{F}uzzy Learning with \underline{P}atch \underline{G}uidance for image ordinal regression, termed \modelname.
Specifically, we first train a \dcl{network} annotator offline using solely the available image-level labels, followed by patch-wise division for patch-level pseudo-label inference. Notably, to leverage the inherent ordinality of labels, we propose an Adjacent Category Mixup (ACM) method that enhances the annotator's discriminative capability between similar samples through controlled mixing of adjacent ordinal categories.}
\yhc{Next, we posit that ordinal label ambiguity stems from two distinct sources: the inherent ambiguity in the features of constituent patch regions, and the fine-grained ambiguity in patch attribute features. To address this dual nature of ambiguity, we propose a Dual-level Fuzzy Learning (DFL) module that quantitatively analyzes label ambiguity through both patch-wise and channel-wise perspectives.}
\qqb{Finally, to further refine the noisy patch-level pseudo-labels,} 
we develop a co-teaching \dcl{strategy} with the
\dcl{Noise-aware Patch Filtering (NPF) paradigm} to reduce the negative impact of noisy and redundant patches on training.
Specifically, based on the co-teaching \dcl{strategy}, this module can employ two identical models that are trained alternately, providing each other with masking matrices for high-confidence patches. 
Our main contributions can be summarized as follows:
\begin{itemize}
    \item \yhc{We propose \modelname, a novel image ordinal regression framework that emulates the human decision-making process by  focusing on discriminative patch-level features for image ordinal regression problem.}
    \item \yhc{We introduce the new ACM mixup method during offline patch label generation and the DFL module to effectively quantify label ambiguity through finer-grained modeling of patch-wise and channel-wise fuzzification.} 
    \item {Extensive experiments across diverse datasets demonstrate the consistent superiority of our \modelname framework compared to state-of-the-art approaches, particularly in detail-sensitive scenarios like DR grading.}
\end{itemize}

\section{Related Work}
\label{sec:formatting}

\subsection{\yhc{Image} Ordinal Regression}
The goal of image ordinal regression is to learn a mapping rule that assigns an input image to a specific rank on an ordinal scale.
\dcl{Numerous methods have tackled the direct ordinal label prediction problem with different approaches to effectively leverage the ordinality.}
One classic approach is $K$-rank~\cite{10.1007/3-540-44795-4_13}, which trained $K-1$ subclassifiers to rank ordinal categories. 
\dcl{Furthermore, some methods~\cite{8099569,7780901} used a series of basic CNNs as $K$-rank classifiers.}
\dcl{Other methods adopted an anchor-based comparison scheme.}
\dcl{For instance, Order Learning~\cite{Lim2020OrderLA} designed a pairwise comparator to classify instance relationships, estimating class labels by comparing input instances with reference instances.
Building on this, Lee and Kim~\cite{lee2021deep} 
developed deep repulsive clustering and order-identity decomposition methods.}
Similarly, MWR~\cite{MWR} leveraged a moving window approach to refine predictions by comparing input images with reference images from adjacent categories.
\dcl{Recently, Ord2Seq~\cite{ord2seq} treated ordinal regression as a sequence prediction process, transforming each ordinal category label into a unique label sequence, inspired by the dichotomous tree structure.}
\dcl{On the other hand, several methods focused on the feature aspect.} 
\dcl{Some methods use probability distributions to model ordinal relationships between labels, enhancing the model's ability to learn representations.}
SORD~\cite{softlabel} converted one-hot labels into soft probability distributions to train an ordinal regressor. 
\dcl{Meanwhile, POEs~\cite{poe} represented each data point as a multivariate Gaussian distribution with an ordinal constraint to capture the inherent characteristics of ordinal regression.}
\dcl{In contrast, some methods focus on data generation to enhance the model's ability to distinguish subtle category differences. For example, CIG~\cite{cigframework} addresses class imbalance and category overlap in image ordinal regression through controllable image generation. OCP-CL~\cite{zheng2024enhancing} disentangles ordinal and non-ordinal content in latent factors, augmenting non-ordinal information to generate diverse images while preserving ordinal content.}
In recent years, with the development of pre-trained VLMs, researchers have explored borrowing the rank concept from the language domain~\cite{NEURIPS2022_e55b3343,numCLIP}.
Unlike previous studies, we explicitly utilize {patch-level} features to uncover key factors influencing human decision-making. \dcl{In addition, we apply fuzzy logic to {quantitatively analyze} the inherent ambiguity and ordinality in the feature-label relationships specific to ordinal regression}. 
\begin{figure*}[th]
  \centering
  \includegraphics[width=0.92\textwidth]{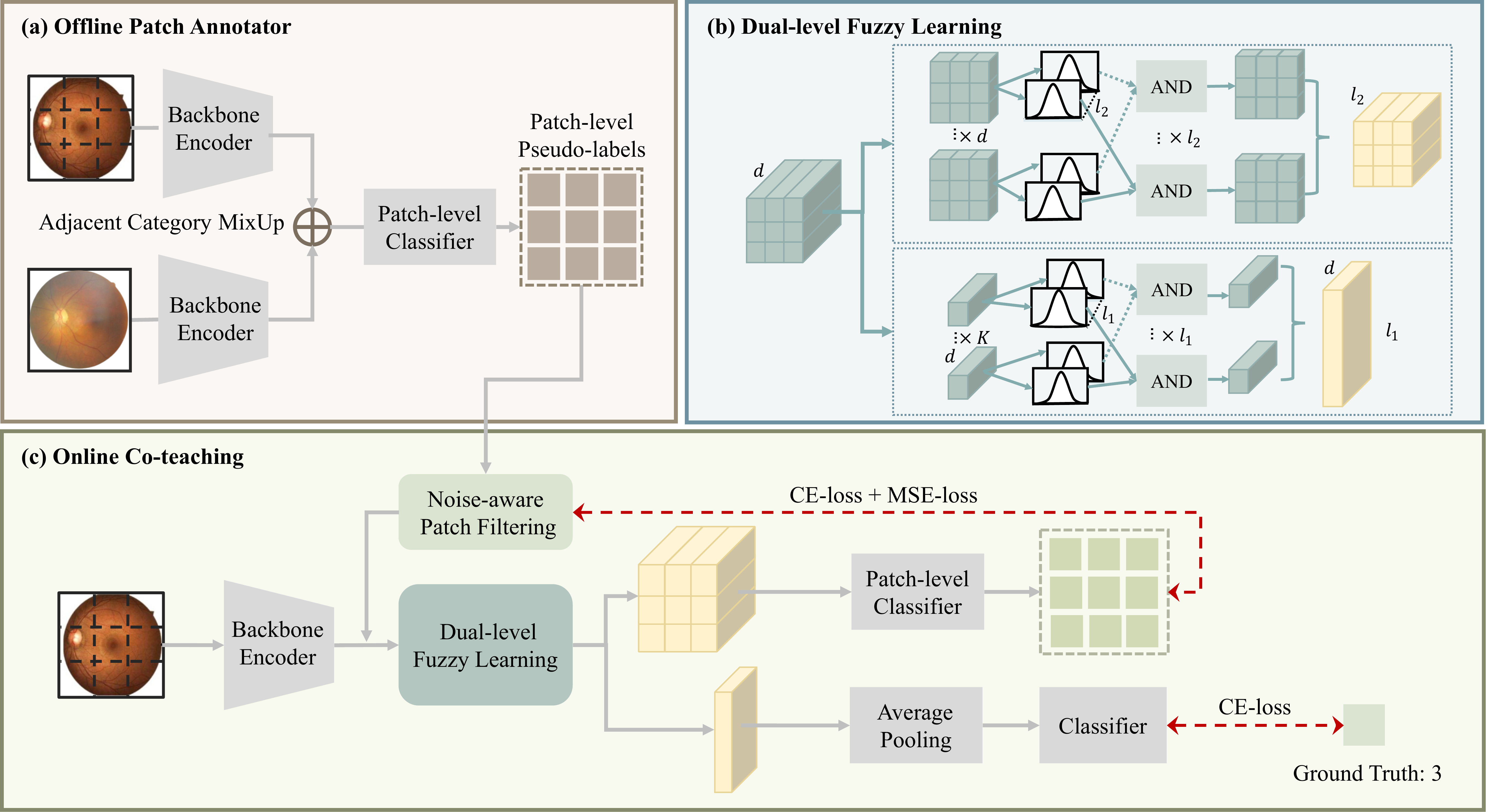} 
  \caption{An overview of our \modelname approach. 
  (a) The Patch Annotator module for generating patch-level pseudo-labels. An adjacent category sampling scheme is adopted to preserve ordinal information inherent in the augmented features, thereby enhancing the model's discriminability on samples of adjacent categories. 
  (b) The Dual-level Fuzzy Learning module, in which multiple Gaussian membership functions are used to introduce fuzziness into the precise image representations, effectively capturing the ambiguity
 in feature-label associations specific to ordinal regression tasks. 
  (c) The overall Co-teaching Strategy of our model, which incorporates a patch-level optimization objective through an unreliable patch filtering method based on the generated pseudo-labels.}
  \label{fig:framework} 
  \vspace{-3ex}
\end{figure*}

\subsection{\dcl{Learning with Label Ambiguity}}
\dcl{The general class distinguishability, human annotator heterogeneity, and external factors can all contribute to the ambiguity in the observed labels, which can impair the model's ability to fit the data.} 
\dcl{Several existing works have tackled this challenge using the label distribution learning (LDL) paradigm~\cite{LDL}, which trained models with instances labeled by label distributions. Furthermore, DLDL~\cite{DLDL} converts each image label into a discrete label distribution and learns the label distribution by minimizing the Kullback-Leibler divergence between the predicted and true label distributions. OLDL~\cite{OLDL} further incorporates modeling of the ordinal nature of labels within the LDL paradigm based on spatial, semantic, and temporal order relationships.
}
\dcl{Another research direction focused on adjusting the representation space based on label ambiguity. In this context, some approaches utilized probabilistic embeddings~\cite{9008376,9157756}, where each sample was represented as a Gaussian distribution rather than a fixed point for classification.}
Furthermore, POEs~\cite{poe} proposed an ordinal distribution constraint to preserve the ordinal relationships in the latent space. 
\dcl{In contrast, our \modelname leverages fuzzy logic to model the ambiguity and ordinality of the category labels simultaneously.}

\section{Methodology}



\subsection{Overview}
\qqb{Unlike typical methods in the field of image ordinal regression, we focus on capturing discriminative patch-level features while fully considering the fuzziness and ordinality of classification labels during the process.}
Let $\mathcal{D}  = \left \{({x}_i, y_i) \in \mathcal{X \times Y} \right \}^N_{i=1} $, \yhc{where $\mathcal{X}$ and $\mathcal{Y}$ represent the sets of images and their corresponding categories, respectively, with $|\mathcal{X}| = |\mathcal{Y}| = N$}. 
\yhc{Each image category $y_i \in \mathcal{C}=\{1, 2, \ldots, C\}$, where $C$ is the total number of categories.}
\qqb{Note that,} there is directionality between different categories, which means that the difference between samples $\{({x}_i, 1),({x}_j, 3)\}$ is bigger than that between samples $\{({x}_i, 1),({x}_k, 2)\}$. This reflects the unique regression characteristics of the problem. The main goal is to obtain a model $F_\theta (\cdot) : \mathcal{X \to Y} $ to accurately predict ordinal labels, consistent with a straightforward classification paradigm. 


\subsection{Offline \qqb{Patch} Annotator}
With the notation given above, for a dataset $\mathcal{D}$, only the global ordinal label of each image is available. This restricted condition makes it challenging to directly model the impact of regional features on decision-making.
\yhc{Thus, we introduce an offline Patch Annotator module to generate the patch-level pseudo-labels, as illustrated in Figure~\ref{fig:framework}(a).}

\textbf{Backbone Selection}.
Given an image sample $(x, y)$, it is first processed through an image encoder to extract feature representations. 
We select the Pyramid Vision Transformer (PVT)~\cite{PVT} as the backbone, which has been shown to be effective for ordinal regression in previous studies~\cite{ord2seq}.
The encoder represents the raw image $x$ as a patch embedding in the latent space, mapping $x \to \mathbf{H} \in \mathbb{R}^{K \times d}$, 
where $K$ is the number of patches and $d$ is the embedding dimension. 
In addition, a fully connected layer and a $1\times 1$ convolution layer are employed as classification heads to process features at different levels. 

\textbf{Patch Annotator}. We first use the global labels to train a simple annotator to generate pseudo-labels for each patch, as shown in Figure~\ref{fig:framework}(a). A simple pre-trained annotator is sufficient to generate 
pseudo-labels at the patch level~\cite{NEURIPS2021_9a49a25d}.
\dcl{Moreover, considering the inherent ordinality of labels, capturing discriminative features that effectively  distinguish adjacent categories becomes challenging without sufficient variability in the training data.}
In this context, augmentation methods such as Manifold Mixup~\cite{pmlr-v97-verma19a} are helpful by creating diverse samples to enhance model performance.
\dcl{To control the modified samples near category boundaries, we introduce an {Adjacent Category Mixup (ACM) scheme.}}
\dcl{Specifically, we first sample adjacent category pairs $({x}_i, {x}_j)$ with ordinal labels $y_i$ and $y_j$, respectively, and $|y_i - y_j| = 1$.} 
The images ${x}_i$ and ${x}_j$ are initially processed through the backbone encoder, mapping them to hidden representations $\mathbf{H}_i$ and $\mathbf{H}_j$. Subsequently, before passing them into the classification layer, the hidden representations are mixed up for augmentation following Manifold Mixup~\cite{pmlr-v97-verma19a}.
The augmented samples $(\tilde{\mathbf{H}}, \tilde{{y}})$ are utilized to train the annotator, which is further used to perform inference on the patch-level hidden representations to generate a patch-level pseudo-label vector 
$\mathbf{c}\in \mathcal{C} ^{K}$ 
for each image.  
\dcl{The annotator, trained with ACM-augmented data, can capture fine-grained discriminative features}, thus enhancing the model's ability to distinguish subtle differences between adjacent category samples.
\dcl{With the supervision of the generated pseudo-labels}, the model can explicitly capture regional features that influence grading labels, aligning with the detailed focus that humans use in decision-making processes.

\subsection{Dual-level Fuzzy Learning} 
%
\dcl{A key aspect of ordinal regression is in exploiting the inherent ordinality among category labels, whose definition, however, is often ambiguous and depends on multi-view features of the images. To address this, we propose a dual-level fuzzy learning module that models the feature-label relationship in a fine-grained manner, capturing both patch-wise and channel-wise interactions to quantify the ambiguity in category boundaries, as illustrated in Figure~\ref{fig:framework}(b).}



\textbf{Patch-\qqb{wise} Fuzzification}.
Based on \qqb{the} fuzzy logic, we first approach the problem from a patch perspective, using fuzzy rules to assess the relationships among patch \qqb{features} at different positions. 
A set of Gaussian membership functions is adopted to fuzzify the $k$-th input patch \qqb{features} $\mathbf{H}^{(k)}$, converting it into corresponding membership grades, which can be formulated as:
\begin{align} \label{equ:2}
g_{mk}(\mathbf{H}^{(k)}) =  e^{-\frac{(\mathbf{H}^{(k)} - \mu_{mk} )^2}{\sigma_{mk}^2}},
\end{align}
where \yhc{$m \in \{1,\dots,\ell_1 \}$ represents the $m$-th rule with $\ell_1$ denoting the total number of rules in the fuzzy system, and $k \in  \{1,\dots,K \}$ represents the $k$-th} patch. In this process, every membership function \dcl{associates the latent representation for each patch} $\mathbf{H}^{(k)}$ with a fuzzy linguistic term label. \dcl{It employs smooth transitions to characterize features at a finer granularity, effectively reducing the ambiguity associated with labels}.
Following this, the \textit{AND} fuzzy logic operation is applied across all the membership grades of $\mathbf{H}$ for a given rule, as formulated below:
\begin{eqnarray} \label{equ:3}
\mathbf{f}^1_m&=&\prod_{k=1}^K g_{mk}(\mathbf{H}^{(k)}).
\end{eqnarray}
Thus, for a given latent representation $\mathbf{H} \in \mathbb{R}^{K \times d}$, we can obtain a set of activation strengths for the fuzzy rules, where each rule is computed by aggregating the membership values of different patches $\mathbf{H}^{(k)}$.
\dcl{These two operations decompose the modeling process of label ambiguity. First, the features are fuzzified by calculating their membership to various fine-grained terms. Then the fuzzy feature-label relationships are learned based on the aggregated fuzzy rules.}
In other words, Equations~(\ref{equ:2}) and (\ref{equ:3}) together form a nonlinear representation projection, mapping $\mathbf{H}\in \mathbb{R}^{K \times d} \to \mathbf{f^1} \in \mathbb{R}^{\ell_1 \times d} $.
The construction of this transformation space aligns with the fuzzy definitions of labels in ordinal regression \dcl{at a patch-level}. Note that, as part of our \modelname, \dcl{these operations} can be regarded as a fuzzy layer, parameterized by a set of Gaussian membership functions, where the mean $\mu$ and variance $\sigma$ are trainable parameters of the shape $\ell_1 \times K$. 
Additionally, these two parameters are consistent for patches at the same spatial position across different images, but vary for patches at different positions within the same image.
\dcl{This branch enhances the model's ability to recognize fuzzified features across different patches.}

\textbf{Channel-\qqb{wise} Fuzzification}. To capture inter-channel relationships within the image attribute representations, we perform fuzzy learning on $\mathbf{H}$ from an alternative perspective, as illustrated in Figure~\ref{fig:framework}(b). This branch is symmetric to that of the patch-level fuzzification, sharing a similar structure. For simplicity, its detailed formalization is omitted here. The difference lies in its implementation of a nonlinear representation projection mapping $\mathbf{H} \in R^{K\times d} \to \mathbf{f}^2 \in R^{K\times \ell_2}$, with trainable parameters of the shape $\ell_2 \times d$. Different channels of the same image are assigned to different membership functions (\ie parameters). This setup is based on the rationale that features within the same channel tend to exhibit similar characteristics in the image representations. 
\dcl{The activation strengths of all the fuzzy rules in this branch measure interactions across channels, enabling the model to extract discriminative channel-wise features.}

\dcl{The final ordinal label prediction is derived collaboratively from the outputs of both branches. Moreover, a $1 \times 1$ convolution layer and a linear layer serve as classification heads for the patch-level and image-level features, respectively, providing the predicted probabilities}.


\subsection{Online Co-teaching Strategy}
Misclassified patch-level pseudo-labels caused by annotator bias introduce erroneous supervisory information during model training. To address this, we propose a 
co-teaching strategy 
to conduct noise-aware patch filtering and provide additional supervision at the patch level, as shown in Figure~\ref{fig:framework}(c).


Different from the prior works on learning with noisy labels, 
which focused on dealing with noise in real-world data and preventing models from overfitting to noisy labels, 
our goal is to actively filter patch-level pseudo-labels generated by the patch annotator. 
Hence, we propose a co-teaching approach~\cite{DivideMix}, which trains \dcl{two versions of the \modelname model}, $F_A$ and $F_B$, simultaneously. Each model assigns reliable (retained) and unreliable (masked) patches to the other’s training dataset based on the patch loss distribution. 
Deep networks have been shown to learn simple and generalizable patterns more quickly than noisy patterns~\cite{10.5555/3305381.3305406}. 
Thus, training samples with smaller loss values are commonly regarded as clean samples.
In the patch filtering method, we divide every training epoch into two steps: Mask Matrix Prediction and Pseudo-label Reflection.

\textbf{Mask Matrix Prediction}. 
A two-component Gaussian Mixture Model (GMM) is initially employed to model the distribution of the cross-entropy loss $\mathcal{L}_{ce}$ across all the patches. By fitting GMM to $\mathcal{L}_{ce}$, it can cluster the patches into two groups based on their loss values. Thus, the credibility probability $w_{k}$ of each patch $x^{(k)}$ can be determined by calculating the posterior probability ${p(g\mid \mathcal{L}_{ce}^{k})}$, where $g$ is the Gaussian component with the smaller mean in GMM. Based on the credibility probability $w_{k}$ and a threshold hyperparameter $\tau$, we construct the patch-level mask matrix $\mathbf{M} \in \{0,1\}^{K}$ for the input image $x$, as:
\begin{eqnarray}
\mathbf{M}_{k} = \left\{\begin{matrix} 1,  ~~~~~ {\rm if} \ w_{k}\geq \tau, \\ 0, ~~~~~{\rm if} \ w_{k}<\tau. \end{matrix}\right.
\end{eqnarray}

\textbf{Pseudo-label Reflection}. Note that directly discarding the masked patch could result in a loss of potentially valuable information and reduce the available ordinal context for learning regional features. Hence, reflection on the pseudo-labels is essential to enhance the model performance. For this, we apply a semi-supervised technique to reprocess the pseudo-labels of both the reliable (retained) and unreliable (masked) patches separately. At each epoch, we train the two models $F_A$ and $F_B$ alternately, keeping one fixed while updating the other. For simplicity, the following description \dcl{takes} model $F_A$ as an example. First, for a reliable patch $x^{(k)}$, its pseudo-label $\mathbf{c}_{k}$ and the model’s new prediction probability $p^A_{k}$ are linearly combined using the corresponding credibility probability $w_{k}$, as:
\begin{eqnarray}
\hat{\mathbf{c}}_{k} = w_{k}\mathbf{c}_{k} + (1-w_{k})p^A_{k}. 
\end{eqnarray}
In contrast, for each unreliable patch, we leverage the average ensemble of the predictions from both models $F_A$ and $F_B$ as the regenerated pseudo-labels with high confidence:
\begin{align} \label{equ::upatch}
\mathbf{c}'_{k} = (p^A_{k} + p^B_{k})/2, \ \ 
\bar{\mathbf{c}}_{k} = \frac{{\mathbf{c}'_{k}}^{1/\delta}}{ \sum\limits_{k \mid \mathbf{M}_{k} = 0} {\mathbf{c}'_{k}}^{1/\delta}},
\end{align}
where $\delta$ is a hyperparameter that sharpens the regenerated probability distribution, making it more concentrated.

\textbf{Optimization}. 
In this study, we tackle ordinal regression under the classification paradigm, which utilizes the traditional Cross-Entropy (CE) objective as the main loss to enhance the classification capacity of the model. It is worth noting that we extend it to be applicable to the reliable pseudo-labels $\hat{\mathbf{c}}$ as an auxiliary loss during the training phase, aiming to leverage regional feature supervision, as:
\begin{align}\label{eq:7}
\mathcal{L}_{cls} =\mathrm{CE} \big(p(x) ,y\big)+ \beta \frac{1}{\| \mathbf{M} \|_1}\sum_{k \mid \mathbf{M}_{k} = 1} \mathrm{CE}\big(p{ (x^{(k)})},\hat{\mathbf{c}}_{k}\big),
\end{align}
where $x^{(k)}$ indicate the $k$-th patch in image $x$. 

In addition, for the regenerated pseudo-labels in Equation~(\ref{equ::upatch}), the Mean Squared Error (MSE) loss is employed:
\begin{align} \label{eq:8}
\mathcal{L}_{re}= \frac{1}{\| \mathbf{1}_{K}-\mathbf{M} \|_1}\sum_{k \mid \mathbf{M}_{k} = 0}\mathrm{MSE} (p(x^{(k)}),\bar{\mathbf{c}}_{k}).
\end{align}
By combining Equations~(\ref{eq:7}) and (\ref{eq:8}), we optimize \modelname through the minimization of the following objective function with a weight hyperparameter $\gamma$.
\begin{align}
\mathcal{L} = \mathcal{L}_{cls} + \gamma  \mathcal{L}_{re},
\end{align}

\section{Experiments}
 In this section, we conduct extensive experiments on datasets under three different scenarios, to evaluate the effectiveness of our proposed \modelname framework.

\subsection{Experimental Setup}
\textbf{Datasets}. First, we utilize the Image Adience dataset~\cite{adience} and the Aesthetics dataset~\cite{Schifanella_Redi_Aiello_2021} to evaluate our approach in general scenarios. 
For the Adience dataset, the 26,580 face images are divided into eight age groups (i.e., the images are labeled from 1 to 8
Similarly, in the Aesthetics dataset, each of the 13,706 images was rated by at least five graders across five ranking categories to assess photographic aesthetic quality. 
The ground truth for each image is determined as the median rank among all the ratings. 
In addition, to demonstrate the broad applicability of \modelname, we employ a Diabetic Retinopathy (DR) dataset in the medical grading domain. The task is to classify fundus images into five levels of diabetic retinopathy, ranging from 1 to 5, 
Note that this dataset is highly imbalanced, with 73.5\% of samples labeled as 0 (for no DR).  
Some statistical information about all the considered datasets is summarized in Table~\ref{tab:dataset}. Detailed descriptions and example images of these datasets can be found in the Appendix.

\textbf{Metrics}. Due to the nature of ordinal regression as an intermediate problem between classification and regression, we evaluate our \modelname from two perspectives. 
First, \dcl{considering the data imbalance in the selected datasets, we find that relying solely on accuracy as a classification metric is insufficient, as accuracy typically reflects the model's performance on the majority categories while overlooking its effectiveness on the minority categories}.
Thus, we incorporate accuracy, precision, recall, and F1-score to provide a more comprehensive evaluation of the model's classification performance. 
Notably, to provide an overall evaluation across all the category labels, we calculate the last three metrics using macro averaging. 
Second, we employ Mean Absolute Error (MAE) between the predicted and ground truth labels, which directly measures the model’s capability to capture ordinal relationships among labels.

\textbf{Implementation Details}. We implement \modelname using PyTorch on an NVIDIA GTX 4090 GPU server. For a fair comparison with existing methods~\cite{cigframework,ord2seq}, we use the PVT architecture as our backbone. The default Adam optimizer~\cite{Adam} is adopted with a batch size of 24, and training is conducted for 50 epochs per stage. We perform 5-fold cross-validation on the Adience and Aesthetics datasets, and 10-fold cross-validation on the DR dataset, reporting the average results.

\begin{table} 
  \centering
  \begin{tabular}{ccc}
    \toprule
    Dataset & \# of Images &Category labels \\
    \midrule
    Adience & 26,580 & 1-8\\
    Aesthetics & 13,706& 1-5\\
    DR & 35,126&1-5\\
    \bottomrule
  \end{tabular}
  \caption{Statistical information of the three evaluation datasets.}
  \label{tab:dataset}
    \vspace{-3ex}
\end{table}


\begin{table*}[t!]
    \belowrulesep=0pt
    \aboverulesep=0pt
    \centering
    \begin{tabular}{cccccccc} 
    \toprule
    Dataset&Metric& CNNPOR & SORD & POE & CIG-PVT & Ord2Seq & \cellcolor{gray!20}\modelname (ours) \\ \hline
    \multirow{5}{*}{Adience} & Precision & - & 0.5430 & 0.5699 
 & {0.5751} & \underline{0.5804} & \cellcolor{gray!20}\textbf{0.5904} \\
                             & Recall & - & 0.5529 & 0.5636 
 & \underline{0.5696} & 0.5668& \cellcolor{gray!20}\textbf{0.5921} \\
                             & F1-score & - & 0.5363 & 0.5580 
 & \underline{0.5678} & 0.5603 & \cellcolor{gray!20}\textbf{0.5801} \\
                             & Accuracy  & 0.5740 & 0.6097 & 0.6159 
 & \underline{0.6288} & 0.6244 & \cellcolor{gray!20}\textbf{0.6360} \\
                             & MAE & 0.5500 & 0.4645 & 0.4713 
 & 0.4429 & \underline{0.4341} & \cellcolor{gray!20} \textbf{0.4327} \\ \hline 

    \multirow{5}{*}{Aesthetics} & Precision & - & 0.4038 & 0.4478 & \textbf{0.4815} & 0.4390 & \cellcolor{gray!20}\underline{0.4512} \\
                             & Recall & - & 0.2773 & 0.3110 & 0.3483 & \underline{0.3484} & \cellcolor{gray!20}\textbf{0.3691} \\
                             & F1-score & - & 0.2885& 0.3285 & \underline{0.3763} & 0.3696 & \cellcolor{gray!20}\textbf{0.3834} \\
                             & Accuracy  & 0.6748 & 0.6875 & 0.6822 & \textbf{0.6988} & 0.6896 & \cellcolor{gray!20}\underline{0.6966} \\
                             & MAE & 0.3540 & 0.5248 & 0.3603 & 0.3340 & \underline{0.3230} & \cellcolor{gray!20}\textbf{0.3187} \\ \hline
    \multirow{5}{*}{\makecell{Diabetic\\ Retinopathy}} & Precision & - & 0.6025 & \underline{0.6244} 
 & {0.6182} & \textbf{0.6294} &  \cellcolor{gray!20}0.6168 \\
                             & Recall & - & 0.4969 & {0.5248} 
 & {0.5088} & \underline{0.5589}  &  \cellcolor{gray!20}\textbf{0.5932} \\
                             & F1-score & - & 0.5241 & 0.5577 
 & {0.5434} & \underline{0.5844}  &  \cellcolor{gray!20}\textbf{0.6008} \\
                             & Accuracy  & 0.8287 & 0.8034 & 0.8285 
 & {0.8303} & \underline{0.8310} &  \cellcolor{gray!20}\textbf{0.8339} \\
                             & MAE & 0.3350 & 0.2865 & 0.2557 
 & 0.3036 & \underline{0.2532} &  \cellcolor{gray!20}\textbf{0.2440} \\ 
    \bottomrule
    \end{tabular}
    \vspace{-1ex}
    \caption{Experimental results on the three evaluation datasets. The best and second-best results are marked in {\bf bold} and \underline{underlined}, respectively. '-' indicates that we could not reproduce the results.}
    \label{tab:comparasion}
    \vspace{-2ex}
\end{table*}

\subsection{Comparison with Known Methods}
To ensure a comprehensive comparison, we \dcl{re-implement} and evaluate several \dcl{cutting-edge} ordinal regression models\dcl{}. Table~\ref{tab:comparasion} summarizes the experimental results of all the methods across all three datasets, from which we draw the following key observations.

Recent ordinal regression works, such as CIG~\cite{cigframework} and Ord2Seq~\cite{ord2seq}, demonstrate promising results across various datasets. Specifically, CIG and Ord2Seq emphasize the critical importance of distinguishing adjacent categories, highlighting the necessity of explicitly exploring the ambiguous boundaries between neighboring category labels.
CIG employs controllable conditional generation to create artificial images based on a base image and its neighboring category samples, which helps the model learn more accurate and robust decision boundaries. Ord2Seq transforms ordinal labels into binary label sequences, using a dichotomy-based sequence prediction method to differentiate adjacent categories through a progressive refinement scheme. \yhc{Their work promotes us to leverage more fine-grained patch features to resolve the ambiguity issue.}

Our model demonstrates superior performance across the three datasets. 
More concretely, on the Adience dataset, our model achieves improvements of {(+1.00\%, +2.25\%, +1.23\%, +0.72\%, 0.0014)} in all the metrics compared to baseline models.
This shows that our model \dcl{more effectively addresses} the challenge of ambiguity in grading boundaries within ordinal regression. Furthermore, the improvements are more pronounced in the precision, recall, and F1-score metrics computed for individual categories. This indicates that our model can effectively distinguish samples across all categories. 
\yhc{Furthermore, our \modelname model also achieves significant improvement on the class-imbalanced datasets, Aesthetics and DR.}
\dcl{Note that, 65.6\% of the samples in Aesthetics dataset are labeled with class 3 (\ie ordinary), and 73\% of the samples in DR dataset are labeled with class 1 (\ie no DR).} 
\yhc{Our model demonstrates performance gains of (+2.07\%, +0.71\%, 0.0043) and (+3.43\%, +1.64\%, 0.0092) in terms of Recall, F1-score, and MAE on Aesthetics and DR, respectively.}
Meanwhile, we attribute the relatively lower performance in Precision and Accuracy on these two datasets to \modelname's enhanced ambiguity modeling ability, which allows it to more effectively learn features of the minority class (see Section~\ref{sec4.3} for details).
Besides, we observe that \modelname achieves the most significant improvement on the DR dataset. This highlights the effectiveness of patch-level supervision for grading decisions, \yhc{aligning with clinical practice where physicians prioritize local lesion assessment for diagnosis.}

\subsection{Minority Class Classification}\label{sec4.3}
\begin{figure*}[t]
	\centering
	\begin{subfigure}{0.246\linewidth}
		\centering
		\includegraphics[width=1\linewidth]{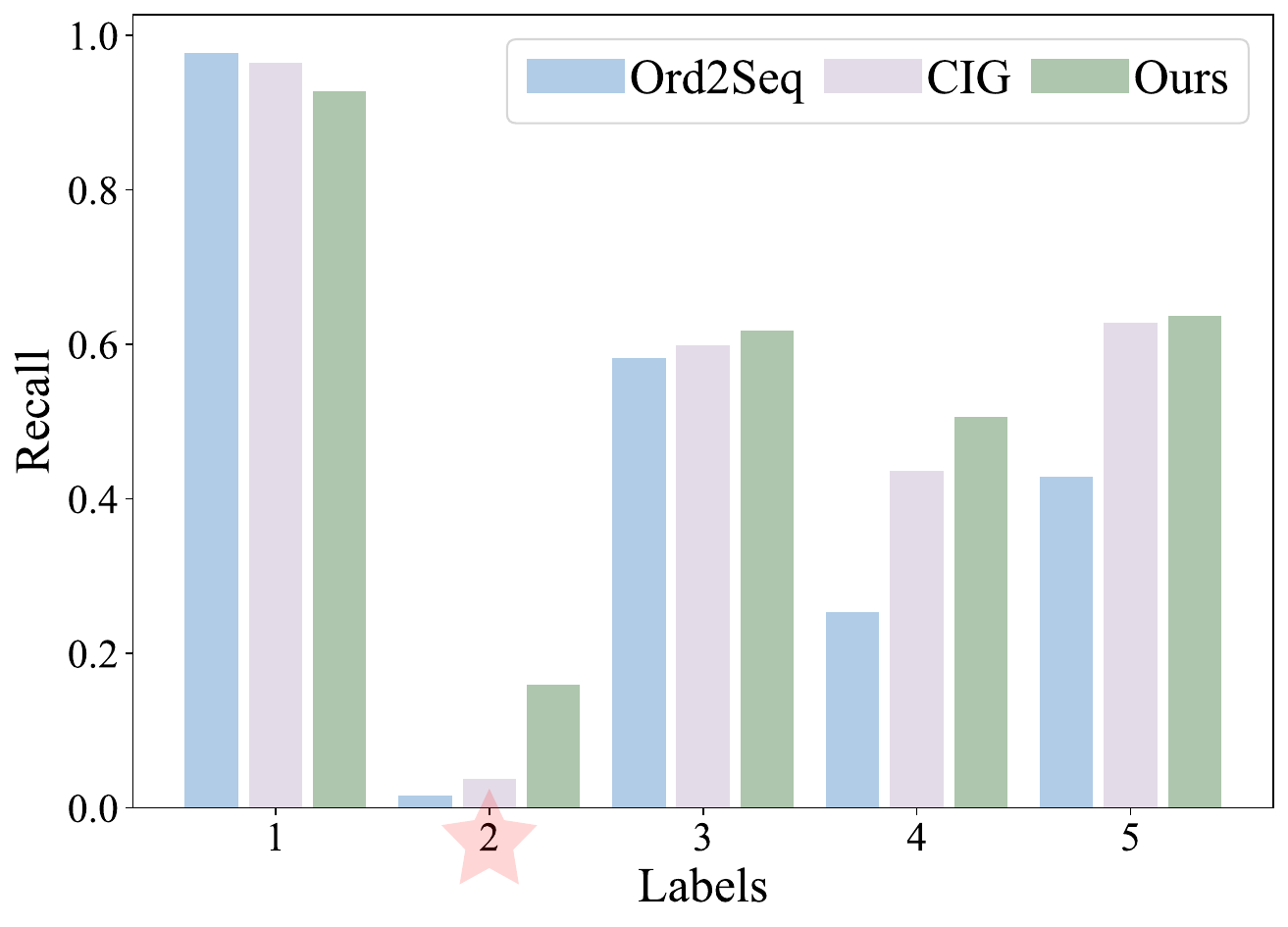}
		\caption{Recall on DR dataset}
		\label{1}
	\end{subfigure}
	\centering
	\begin{subfigure}{0.246\linewidth}
		\centering
		\includegraphics[width=1\linewidth]{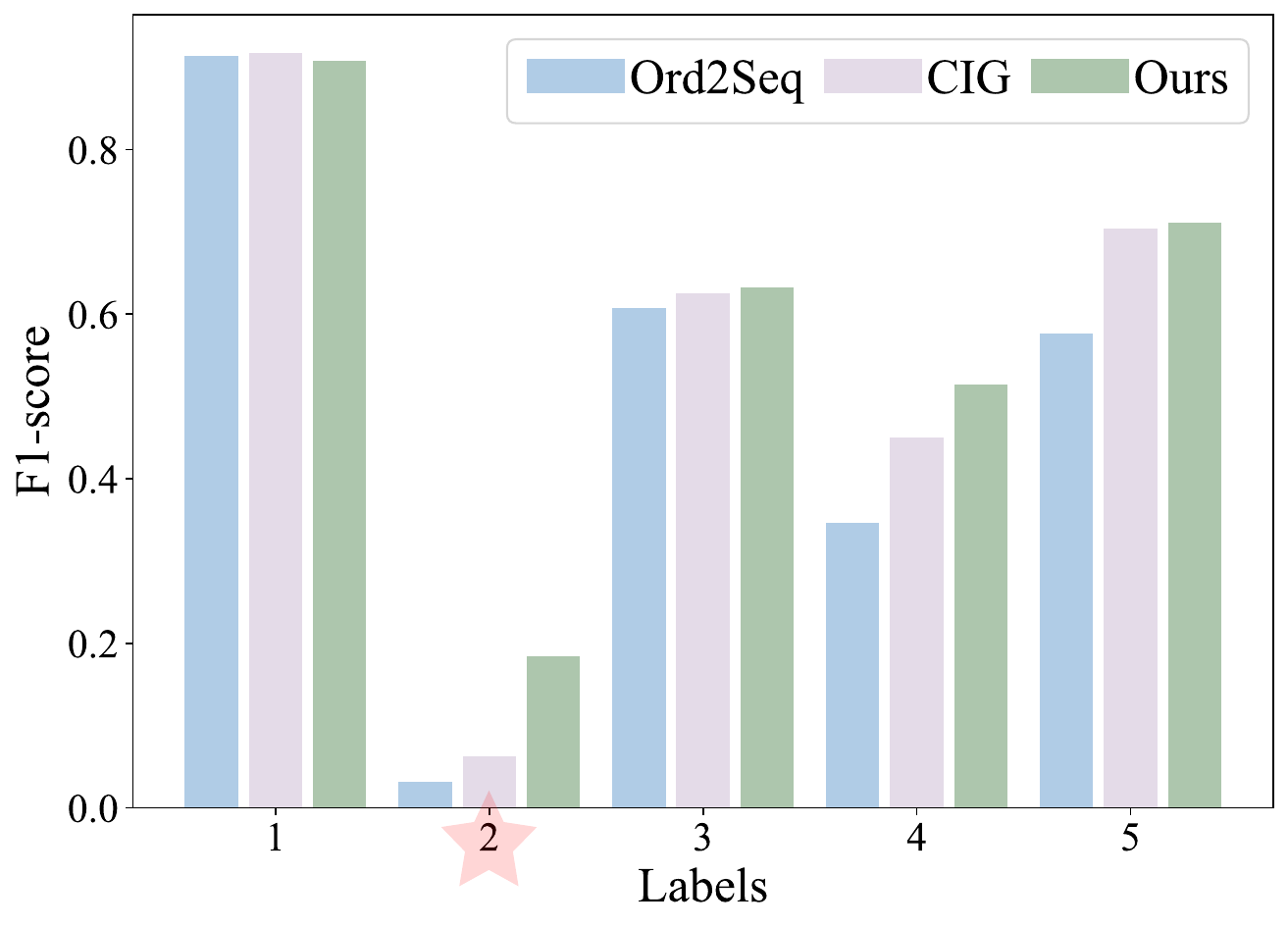}
		\caption{F1-score on DR dataset}
		\label{2}
	\end{subfigure}
	\begin{subfigure}{0.246\linewidth}
		\centering
		\includegraphics[width=1\linewidth]{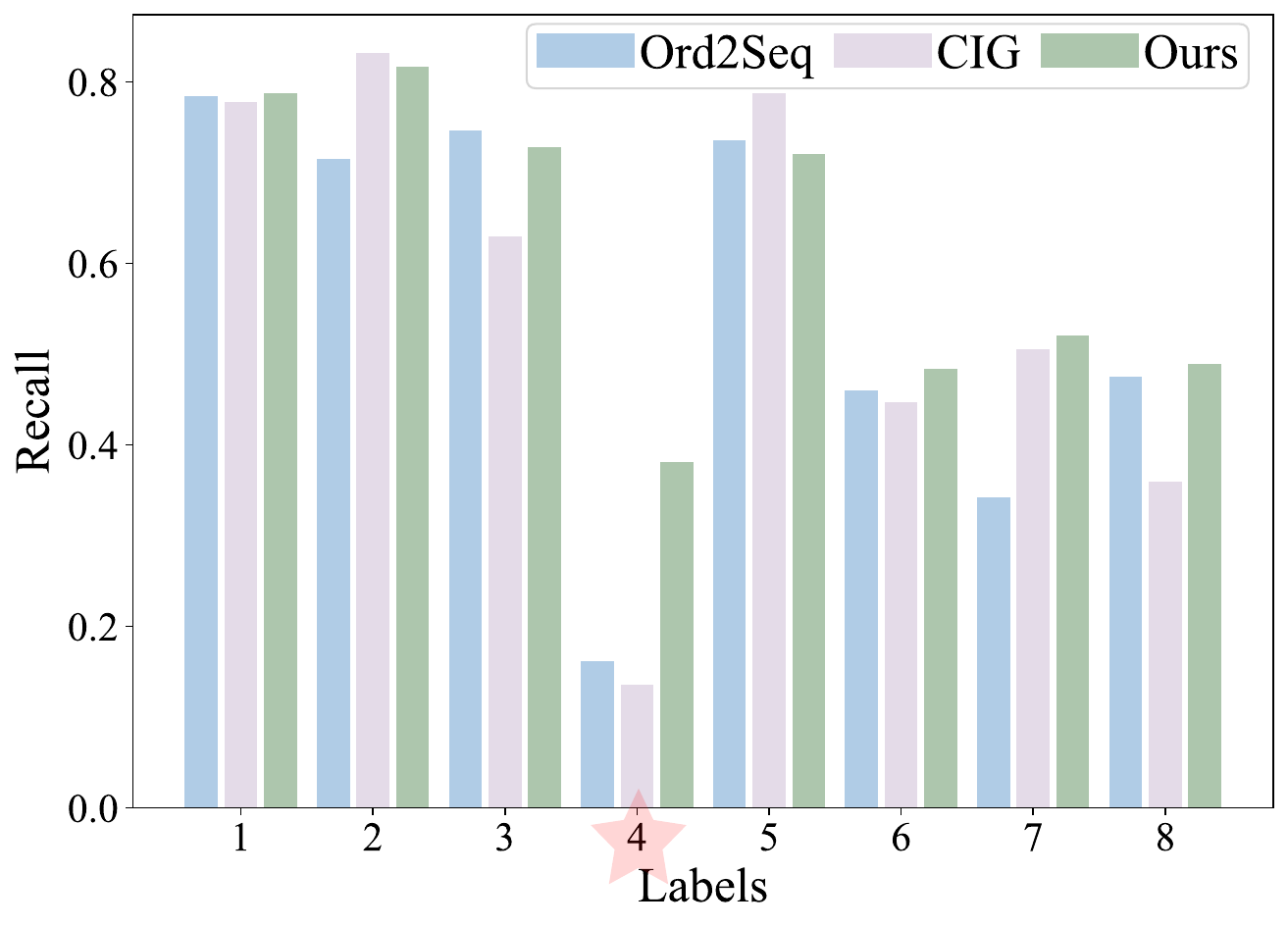}
		\caption{Recall on Adience dataset}
		\label{3}
	\end{subfigure}
 	\begin{subfigure}{0.246\linewidth}
		\centering
		\includegraphics[width=1\linewidth]{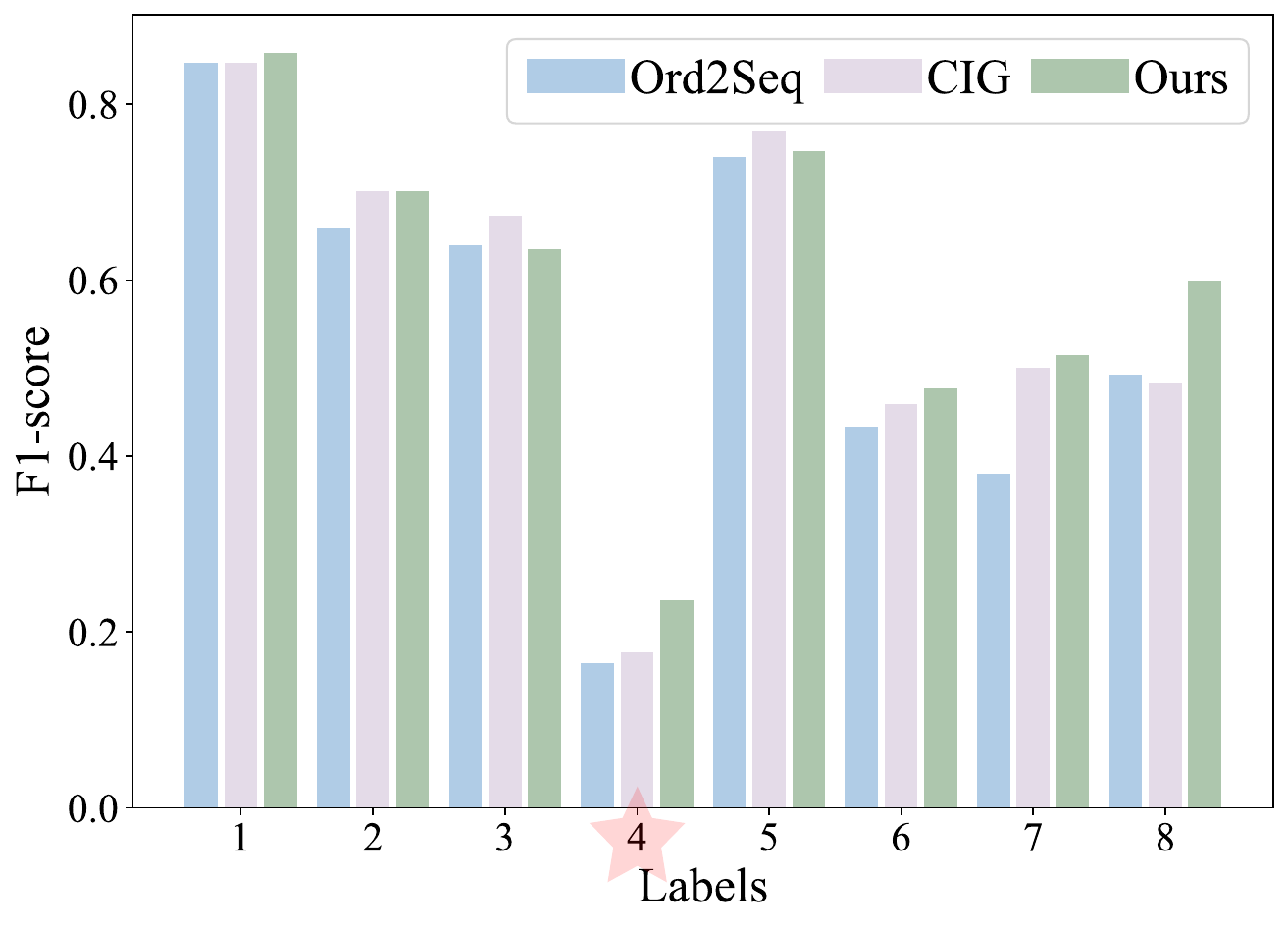}
		\caption{F1-score on Adience dataset}
		\label{4}
	\end{subfigure}
    \vspace{-4ex}
    \caption{Detailed performance for each category on the DR and Adience datasets. 
    We show two evaluation metrics, Recall and F1-score. The star symbol indicates the minority category.}
  \label{fig:Minority} 
  \vspace{-3ex}
\end{figure*}

A typical classification framework using the Cross-Entropy loss often suffers from frequent ``passive updates" of minority classes, resulting in low separability among these classes. 
Hence, we conduct evaluations for each ordinal class on the DR and Adience datasets to examine the robustness and effectiveness of our model under different data distributions. From the results in Table~\ref{tab:comparasion}, we compare our model with two state-of-the-art methods, Ord2Seq and CIG\dcl{, which achieved the best performance in their respective baseline methods}. 
Figure~\ref{fig:Minority} presents the detailed evaluation results for each class with these two methods. \yhc{The corresponding similar observation on the Aesthetics dataset can be found in the Appendix.}

On both datasets, there are typical minority or hard-to-distinguish categories, such as class 2 in the DR dataset and class 4 in the Adience dataset. The performance of the three models (Ord2Seq, CIG-PVT, and our \modelname) shows a notable decline in these categories. For example, in level 2 of the DR dataset, the Recall values of the three models are only (3.69\%, 1.64\%, and 15.98)\%, respectively, which is significantly lower than their average recall on this dataset {(50.88\%, 55.89\%, and 59.32\%)}. This is because the limited number of minority category samples restricts the model's ability to effectively capture distinguishing features. 
This underscores the importance of leveraging features from adjacent categories to aid in classifying minority categories is a valuable research direction for image ordinal regression.

\begin{table}[t!]
  \centering
      \belowrulesep=0pt
    \aboverulesep=0pt
  \begin{tabular}{ccc|ccc}
    \toprule
    \multicolumn{3}{c|}{Method}&\multicolumn{3}{c}{Diabetic Retinopathy} \\ 
    DFL& PA& NPF&F1-score&Acc&MAE\\
    \midrule
     -&-&-&0.5368&0.8261&0.2636\\ 
     \checkmark&-&-&0.5808&0.8285&0.2524\\
    \checkmark& \checkmark&-&0.5838&0.8311&0.2485\\
     \rowcolor{gray!20}\checkmark& \checkmark& \checkmark&
     \textbf{0.6008}&\textbf{0.8339}&\textbf{0.2440}\\

    \bottomrule
  \end{tabular}
  \vspace{-1ex}
  \caption{Results of ablation study on the DR dataset.}
  \label{tab:abl}
    \vspace{-2ex}
\end{table}



\begin{table}[t]
    \belowrulesep=0pt
    \aboverulesep=0pt
  \centering
  \begin{tabular}{c|cccc}
    \toprule
      Method&Recall&F1-score&Acc&MAE\\
    \midrule
     Mixup &0.5797&0.5682&0.6308&0.4442 \\
     \rowcolor{gray!20}ACM &\textbf{0.5921}&\textbf{0.5801}&\textbf{0.6360}&\textbf{0.4427}\\
    \bottomrule
  \end{tabular}
  \vspace{-1ex}
  \caption{Results of ablation study for the Mixup method on the Adience dataset.}
  \label{tab:abl_mixup}
    \vspace{-4ex}
\end{table}

However, despite a performance decline, our model still outperforms the two counterparts in the minority classes. 
While CIG-PVT leverages controllable generation techniques to supplement minority class samples to improve its performance on these classes, 
our model enhances minority class recognition by utilizing diverse information from neighboring class samples through DFL module.  
Consequently, on the DR dataset, our model achieves a notable improvement of (+12.30\%, +12.23\%) in recall and F1-score for class 2. This is a substantial improvement, as these metrics for this class in the baseline models are typically around 3\% and 5\%. 
We attribute this improvement to two factors. First, the limited samples of class 2 in the DR dataset (only 7\%) make it challenging for the model to capture distinctive features. Second, the high proportion of adjacent-level samples (73.5\% in level 1) and the ordinality of labels causes the model to favor more prevalent neighboring class. 
\dcl{This highlights the superiority of \modelname to distinguish minority category samples by leveraging fine-grained, order-related features from adjacent samples in the membership-based latent space.}

Similarly, our model achieves an improvement of 21.93\% in recall and 5.88\% in F1-score for class 4 on the Adience dataset. Additionally, across most categories in both datasets, our model shows superior performance. 
This further demonstrates the robustness of our model, as it consistently enhances 
performance across various image datasets.

%


\subsection{Ablations}
We conduct ablation studies to empirically verify the rationality of \modelname design. We evaluate three main components of the framework, \ie dual-level fuzzy learning (DFL), patch annotator (PA), and noise-aware patch filtering (NPF). 
Due to page limitations, we present an analysis based solely on the metrics in Table~\ref{tab:abl} for the DR dataset, 
with analyses of the other two datasets provided in the Appendix.

We consider the PVT network with a linear layer as the fundamental framework, and conduct experiments based on this structure. The results show that our dual-level fuzzy learning (DFL) module is highly effective. By applying fuzzification to the image representations through DFL alone, we improve F1-score, Accuracy, and MAE by (+4.35\%, +0.24\%, 0.011), respectively. These results indicate that addressing ordinal regression with a traditional classification paradigm is insufficient, 
and DFL benefits the model in assessing the ordinal label ambiguity.
Moreover, with the introduction of additional patch-level supervision, the model achieves improvements of {(+0.3\%, +0.26\%, 0.0112)} in the three metrics. In comparison, after filtering the patch-level pseudo-labels, the model achieves improvements of (+2\%, +0.54\%, 0.0084).
This highlights that \dcl{the filtered patch-level features are effective for grading decisions}, while using pseudo-labels from a simple offline annotator yields only limited improvement. 


We further validate the effectiveness of our Adjacent Category Mixup (ACM) strategy. Table~\ref{tab:abl_mixup} presents the results on the Adience dataset, while results for the other datasets are provided in the Appendix. 
Compared to the Manifold Mixup~\cite{pmlr-v97-verma19a}, when using the pseudo-labels generated by the annotator trained with ACM-augmented data as supervision, \modelname shows improvements across all evaluation metrics.
This indicates that ACM method augments samples near category boundaries, helping the model learn distinguishing features between adjacent levels.


\section{Conclusions}
In this paper, \dcl{we presented a novel Dual-level Fuzzy Learning with Patch Guidance (\modelname) framework for image ordinal regression focusing on discriminative patch-level features with only available image-level labels.}
First, we incorporated the patch annotator and noise-aware filtering paradigm to leverage only image-level labels for learning informative patch-level feature. 
In this approach, the model emulated the human decision-making process by placing additional focus on discriminative patch-level features to make the final prediction.
\dcl{To explore the ordinal label ambiguity, we further developed the dual-level fuzzy learning module that modeled ambiguous feature-label relationships via fuzzy rule embeddings from both patch-wise and channel-wise perspectives.} 
Extensive experimental results on three different image ordinal regression datasets demonstrated the superiority of \modelname compared to state-of-the-art methods. Additionally, we conducted detailed metric evaluations on specific categories to further illustrate the robustness and effectiveness of \modelname.

\section*{Acknowledgments}
This research was partially supported by National Natural Science Foundation of China under Grant No.\thinspace 62476246 and No.\thinspace 92259202, ``Pioneer" and ``Leading Goose" R\&D Program of Zhejiang under Grant No.\thinspace 2025C02132, and GuangZhou City’s Key R\&D Program of China under Grant No.\thinspace 2024B01J1301.

\bibliographystyle{named}
\bibliography{ijcai25}

\newpage
\appendix

\begin{figure}[t]
  \centering
  \includegraphics[width=0.48\textwidth]{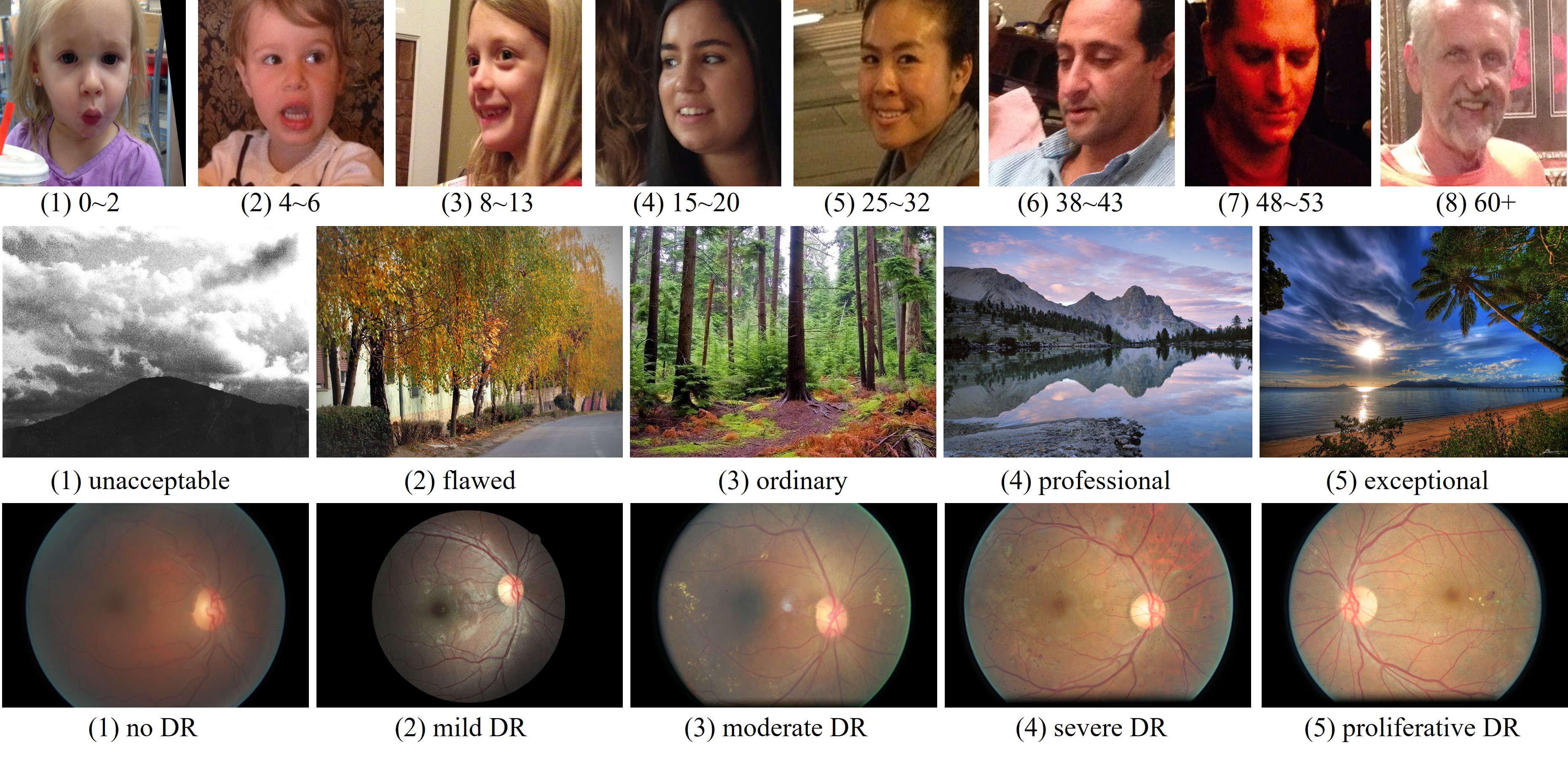} 
  \vspace{-2ex}
  \caption{Datasets visualization. Examples of each of the ordinal categories in three datasets.}
  \label{fig:data_exp} 
  \vspace{-3ex}
\end{figure}

\section{Dataset Details} \label{dataset_detail}
In this section, we will provide a more detailed introduction of the three datasets used. Figure~\ref{fig:data_exp} illustrates the ordinal categories in all datasets with example images. 

\textbf{Adience Dataset} is a face image dataset from Flickr and contains 26,580 face images of 2,284 subjects, which are divided into 8 ordinal categories:  0–2, 4–6, 8–13, 15–20, 25–32, 38–43, 48–53, and over 60 years. In our experiment, all the images are divided into five subject-exclusive folds for cross-validation.

\textbf{Aesthetics dataset} is another Flickr image dataset that contains 15,687 Flickr image URLs, 13,706 of which are available.  These images were rated on a scale from 1 to 5 by at least five evaluators based on image aesthetic quality, with the ground truth defined as the median rank of all the ratings. The five ratings correspond to the following semantic categories: unacceptable, flawed, ordinary, professional, and exceptional. Additionally, in the experiment, we treat this dataset as imbalanced, with the specific number of samples in each category as follows: 248 (1.81\%), 3,315 (24.19\%), 9,002 (65.68\%), 1,116 (8.14\%), and 25 (0.18\%). Following the previous works, we also apply five-fold cross-validation to this dataset.

\textbf{Diabetic Retinopathy} dataset contains 35,126 high-resolution fundus images of patients, classified according to the severity of retinal lesions. Specifically, the images are labeled into five levels: no DR (25,810 images), mild DR (2,443 images), moderate DR (5,292 images), severe DR (873 images), and proliferative DR (708 images), respectively. We selected this dataset to validate the effectiveness and robustness of the proposed model in an imbalanced and specialized application scenario. We apply subject-independent ten-fold cross-validation to this dataset as in previous works.
 \begin{figure}[t]
	\centering
	\begin{subfigure}{0.48\linewidth}
		\centering
		\includegraphics[width=0.98\linewidth]{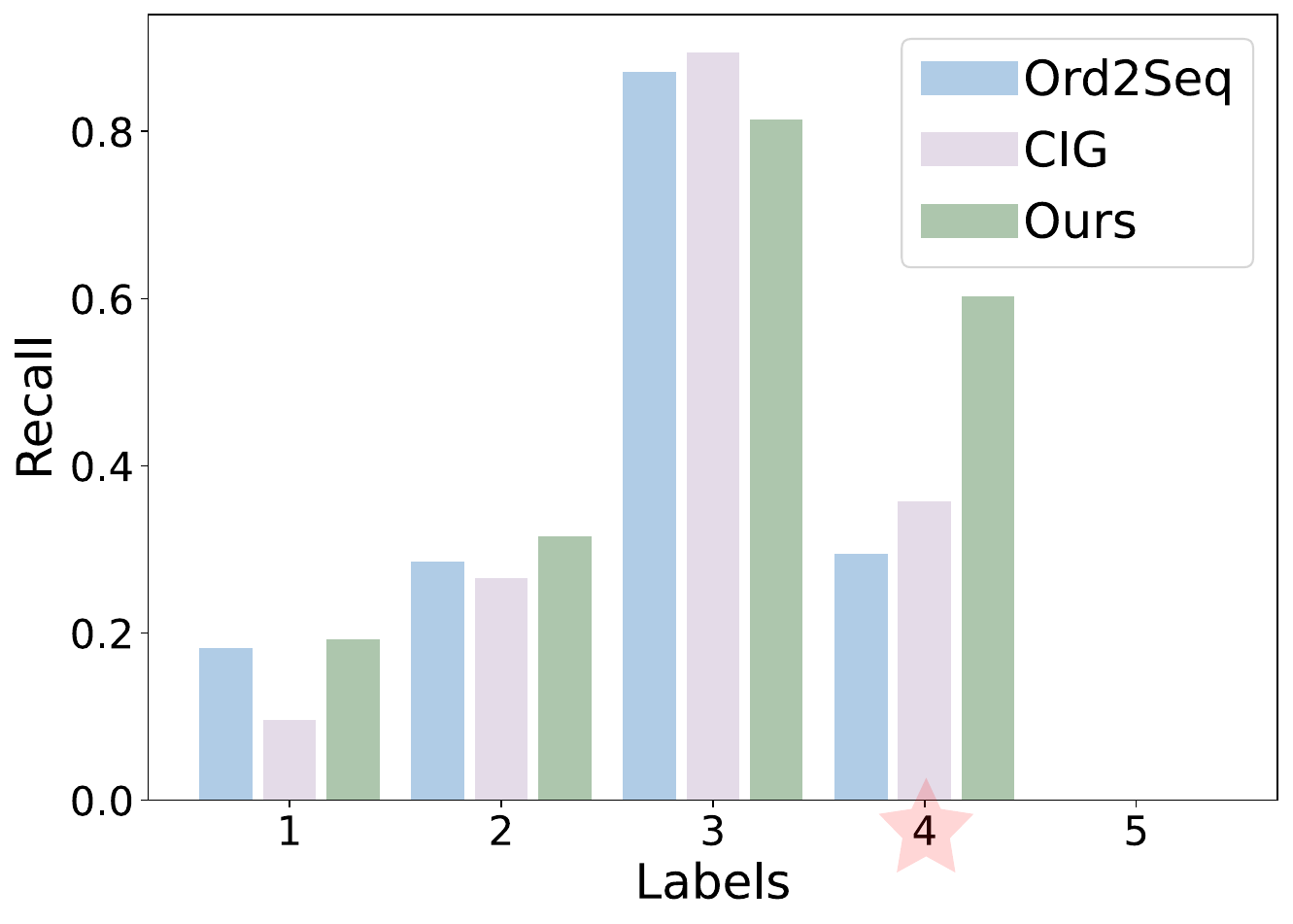}
		\caption{Recall on Aesthetics}
		\label{1}
	\end{subfigure}
	\centering
	\begin{subfigure}{0.48\linewidth}
		\centering
		\includegraphics[width=0.98\linewidth]{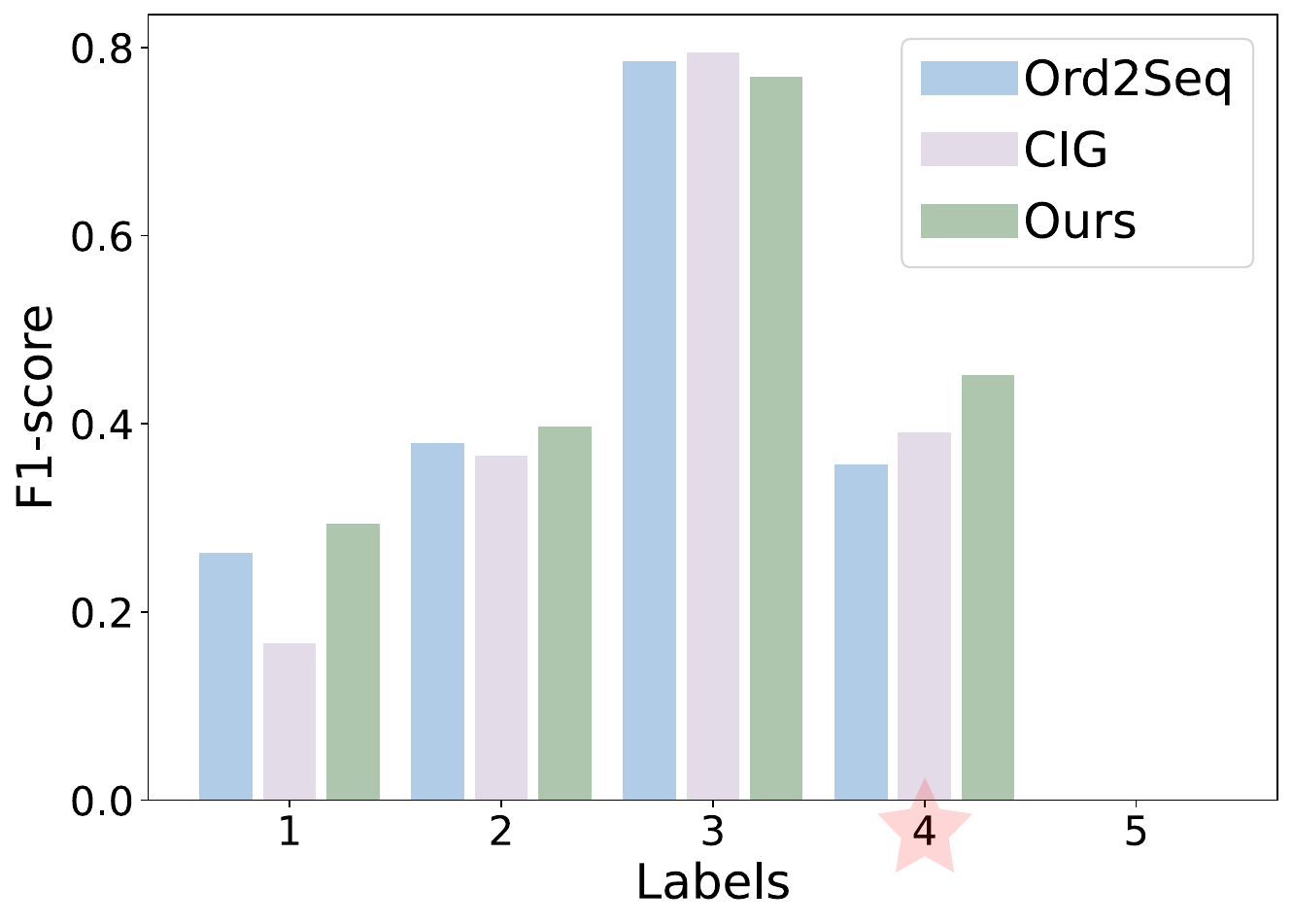}
		\caption{F1-score on Aesthetics}
		\label{2}
	\end{subfigure}
    \caption{Detailed performance for each category on the Aesthetics datasets. 
    We show two evaluation metrics, Recall and F1-score. The star symbol indicates the minority category.}
  \label{fig:Aes_minority} 
\end{figure}

\section{Minority Class Classification for Aesthetics dataset}
In this chapter, we present the details of the minority class classification in the Aesthetics dataset. Following the experiments in the main text, we conduct experiments with Ord2Seq, CIG, and our \modelname, and the detailed evaluation results for each category is shown in the Figure~\ref{fig:Aes_minority}.

For the Aesthetics dataset, the specific class proportions are as described in the Section~\ref{dataset_detail}. This is an extremely imbalanced dataset, where the class with the fewest samples accounts for only 25/13706,~\textit{i.e.}~0.18\% of the total. As a result, all models fail to identify the features of this class (rating 5) and classify it accurately. Therefore, we analyze the results for the other classes and observe that \modelname shows improvements in classes 1, 2, and 4. Notably, the most significant improvement is seen in class 4, which is adjacent to the most common class 3. This result is consistent with the analysis presented in the main text for the Adience and DR datasets, suggesting that \modelname effectively leverages available data to capture discriminative features for minority class samples.

\section{More Ablations}
\label{sec:more_ablations}
We present here some of the analyses and results of the ablation experiments that were omitted from the main text due to space constraints.

\begin{table*}[t!]
  \centering
      \belowrulesep=0pt
    \aboverulesep=0pt
  \begin{tabular}{ccc|ccc|ccc|ccc}
    \toprule
    \multicolumn{3}{c|}{Method}&\multicolumn{3}{c|}{Adience}&\multicolumn{3}{c|}{Aesthetics}&\multicolumn{3}{c}{Diabetic Retinopathy} \\ 
    DFL& PA& NPF&F1-score&Acc&MAE&F1-score&Acc&MAE&F1-score&Acc&MAE\\
    \midrule
     -&-&-&0.5508&0.6204&0.4640&0.3671&0.6862&0.3521&0.5368&0.8261&0.2636\\ 
     \checkmark&-&-&0.5588&0.6214&0.4635&0.3765&0.6906&0.3230&0.5808&0.8285&0.2524\\
    \checkmark& \checkmark&-&0.5645&0.6264&0.4474&0.3765&0.6936&0.3206&0.5838&0.8311&0.2485\\
     \rowcolor{gray!20}\checkmark& \checkmark& \checkmark&
     \textbf{0.5801}&\textbf{0.6360}&\textbf{0.4427}&\textbf{0.3834}&\textbf{0.6966}&\textbf{0.3187}&\textbf{0.6008}&\textbf{0.8339}&\textbf{0.2440}\\

    \bottomrule
  \end{tabular}
  \caption{Results of ablation study on the three datasets.}
  \label{tab:abl}
    \vspace{-2ex}
\end{table*}
\begin{table*}[t]
    \belowrulesep=0pt
    \aboverulesep=0pt
  \centering
  \begin{tabular}{c|cccc|cccc}
    \toprule
    &\multicolumn{4}{c|}{Aesthetics}&\multicolumn{4}{c}{Diabetic Retinopathy} \\ 
    &Recall&F1-score&Acc&MAE&Recall&F1-score&Acc&MAE\\

    \midrule
     Mixup &0.3566&0.3815&0.6915&0.3213 &0.5763 & 0.5847& 0.8295&0.2504 \\
     \rowcolor{gray!20}ACM  
&\textbf{0.3691}&\textbf{0.3834}&\textbf{0.6966}&\textbf{0.3187}
     &\textbf{0.5932}&\textbf{0.6008}&\textbf{0.8339}&\textbf{0.2440}\\
    \bottomrule
  \end{tabular}
  \caption{Results of ablation study for the Mixup method in the Aesthetics and DR datasets.}
  \label{tab:abl_mixup_ap}
    \vspace{-2ex}
\end{table*}

\textbf{Components ablation} analyses in Adience and Aesthetics datasets. The detailed performance of each component ablation is presented in Table\ref{tab:abl}. 
In the Adience dataset, the performance trends observed after component ablation are identical to those analyzed in the main text for the DR dataset. When using only the dual-level fuzzy learning (DFL) module, improvements of (+0.80\%, +0.10\%, and 0.0005) are achieved in F1-score, accuracy, and MAE, respectively. 
Similarly, when the patch annotator (PA) and the noise-aware patch-level filtering (NPF) module are introduced sequentially, we observe improvements of (+0.57\%, +0.50\%, 0.0161) after introducing the PA, and further improvements of (+1.56\%, +0.96\%, 0.0047) after adding the NPF module, across the three metrics.
However, for the Aesthetics dataset, the performance trend differs slightly: compared to the full model, introducing only the patch-level pseudo-label supervision without filtering results in no improvement in the F1-score. This further underscores the importance of the proposed NPF module.

\textbf{ACM strategy ablation} in Aesthetics and DR datasets. The results of the experiment are presented in Table~\ref{tab:abl_mixup_ap}. The consistent improvements across all metrics highlight the effectiveness of the Adjacent Category Mixup (ACM) method we employed.

\section{Experimental details}
We investigate the impact of some hyperparameters in \modelname and present the results on the Adience dataset in Table~\ref{tab:hyper}. 
Specifically, \modelname achieves the best performance with $\tau=0.9,~\beta=1$ and $\ell=256$.
For the hyperparameters $\delta$ and $\gamma$ used in the Co-teaching strategy, we follow the settings from the existing study DivideMix. It sets $\delta=0.5$, while $\gamma$ increases linearly with the training epochs, ranging from $1e^{-5}$ to $150$. The effectiveness of these hyperparameters is validated on a general image dataset. Furthermore, for the size of the patch, we reshape the input image to $224\times224$ and divide it into patches of size $32\times32$, resulting in patches $K=7\times7=49$. 

\begin{table}[t]
  \centering
  \begin{tabular}{cc|cccc}
    \toprule
      \multicolumn{2}{c}{Hyperparameter}&Recall&F1-score&Acc&MAE\\
    \midrule
    \multirow{3}{*}{$~~~\tau$}& $0.6$ &0.5801
 & 0.5628 &0.6236 & 0.4548 \\ & $0.8$ &0.5863 & 0.5751&0.6295 &0.4457\\& $0.9$ &0.5921 &0.5801 & 0.6360&0.4327\\
    \midrule
     \multirow{3}{*}{ $~~~\beta$}& $0.5$ & 0.5818& 0.5667& 0.6255& 0.4503\\ & $1.0$ & 0.5958& 0.5778& 0.6338&0.4352\\& $2.0$ & 0.5908& 0.5682& 0.6312&0.4430\\
     \midrule
     \multirow{3}{*}{ $~~~\ell$}& $64$ & 0.5858 & 0.5793& 0.6255&0.4570 \\ & $256$ &0.5902 & 0.5803& 0.6318&0.4412\\& $512$ & 0.5876 & 0.5788& 0.6297&0.4537\\

    \bottomrule
  \end{tabular}
  \vspace{-1ex}
  \caption{Performance of \modelname with different hyperparameter values on Adience dataset.}
  \label{tab:hyper}
    \vspace{-3ex}
\end{table}

\end{document}